\documentclass[letterpaper, 10 pt, journal, twoside]{IEEEtran}  

\usepackage[pdftex]{graphicx}

\IEEEoverridecommandlockouts                              

\usepackage{amsmath}
\usepackage{booktabs, makecell, multirow, rotating}
\usepackage{graphicx}
\usepackage{float}
\usepackage{caption}
\usepackage{cite}
\usepackage{textcomp}
\usepackage{subfigure}
\usepackage{multicol, blindtext}
\usepackage{tabularx}
\usepackage{hyperref}
\usepackage[dvipsnames]{xcolor}
\title{\LARGE \bf
G.O.G: A Versatile Gripper-On-Gripper Design for Bimanual Cloth Manipulation with a Single Robotic Arm
}

\author{Dongmyoung~Lee$^{*1}$,~Wei~Chen$^{*1}$,~Xiaoshuai~Chen$^{1}$, and~Nicolas~Rojas$^{1}$,~\IEEEmembership{Member,~IEEE}
\thanks{Manuscript received: October 26, 2023; Revised: December 20, 2023; Accepted: January 15, 2024.}
\thanks{This paper was recommended for publication by Editor Júlia Borràs Sol upon evaluation of the Associate Editor and Reviewers' comments. Wei Chen was supported in part by the China Scholarship Council and the Dyson School of Design Engineering, Imperial College London.}
\thanks{$^{1}$Dongmyoung Lee, Wei Chen, Xiaoshuai Chen, and Nicolas Rojas are with the REDS Lab, Dyson School of Design Engineering, Imperial College London, 25 Exhibition Road, London, SW7 2DB, UK
{\tt\footnotesize (d.lee20, w.chen21, c.xiaoshuai19, n.rojas)@imperial.ac.uk}}
\thanks{$^{*}$These authors contributed equally to this work.}
}

\begin{document}

\maketitle





\begin{abstract}
The manipulation of garments poses research challenges due to their deformable nature and the extensive variability in shapes and sizes. Despite numerous attempts by researchers to address these via approaches involving robot perception and control, there has been a relatively limited interest in resolving it through the co-development of robot hardware. Consequently, the majority of studies employ off-the-shelf grippers in conjunction with dual robot arms to enable bimanual manipulation and high dexterity. However, this dual-arm system increases the overall cost of the robotic system as well as its control complexity in order to tackle robot collisions and other robot coordination issues. As an alternative approach, we propose to enable bimanual cloth manipulation using a single robot arm via novel end effector design---sharing dexterity skills between manipulator and gripper rather than relying entirely on robot arm coordination. To this end, we introduce a new gripper, called {\it G.O.G.}, based on a gripper-on-gripper structure where the first gripper independently regulates the span, up to 500mm, between its fingers which are in turn also grippers. These finger grippers consist of a variable friction module that enables two grasping modes:  firm and sliding grasps. 
Household item and cloth object benchmarks are employed to evaluate the performance of the proposed design, encompassing both experiments on the gripper design itself and on cloth manipulation. Experimental results demonstrate the potential of the introduced ideas to undertake a range of bimanual cloth manipulation tasks with a single robot arm. Supplementary material is available at \href{https://sites.google.com/view/gripperongripper}{https://sites.google.com/view/gripperongripper}.
\end{abstract}

\begin{IEEEkeywords}
Bimanual cloth manipulation, single robot arm, gripper design. 
\end{IEEEkeywords}

\section{Introduction}
\IEEEPARstart{T}{he} robotic manipulation of clothes and fabrics has been a topic of great interest among researchers for a long time, but it comes with a variety of challenges.
These can be fundamentally attributed to the deformable nature of garments in real-life situations, including factory and domestic environments~\cite{borras2022effective}. These deformable characteristics make tasks such as perception, grasping, and manipulation inherently challenging for a robot.
Regarding cloth perception, notable advancements have been observed in recent studies, particularly in areas such as state estimation and grasping point detection~\cite{chi2021garmentnets,ren2023autonomous,10342086,sanchez2018robotic}. 
Regarding cloth grasping and manipulation, the majority of studies enables bimanual manipulation utilizing two robot arms~\cite{sahari2010clothes, miller2012geometric, doumanoglou2016folding, weng2022fabricflownet,ha2022flingbot, sunil2023visuotactile}. This natural approach introduces a variety of challenges though, encompassing collision avoidance, intricate control strategies, and larger costs.

In response to these challenges, some researchers have put effort into the domain of single-armed cloth manipulation by refining the hardware of the gripper itself. 
For instance, several grippers have been proposed to facilitate intricate fabric manipulation~\cite{sugiura2010foldy, von2017naist}. However, these designs exhibit limitations in terms of versatility, either (i) failing to accommodate fabrics of varying dimensions or (ii) struggling to execute diverse tasks, such as folding and flattening, especially when utilizing a single robot arm. A recent development entails a multi-fingered gripper~\cite{donaire2020versatile} with the potential for addressing multiple cloth manipulation tasks, although its integration with a robot arm for automated execution can present considerable complexities. There is then an opportunity to augment the capabilities of end-effectors for these tasks, specifically in the context of single-armed bimanual cloth manipulation.

\begin{figure}[t!]
    \centering
    \vspace{4mm}
    \includegraphics[width=0.95\columnwidth]{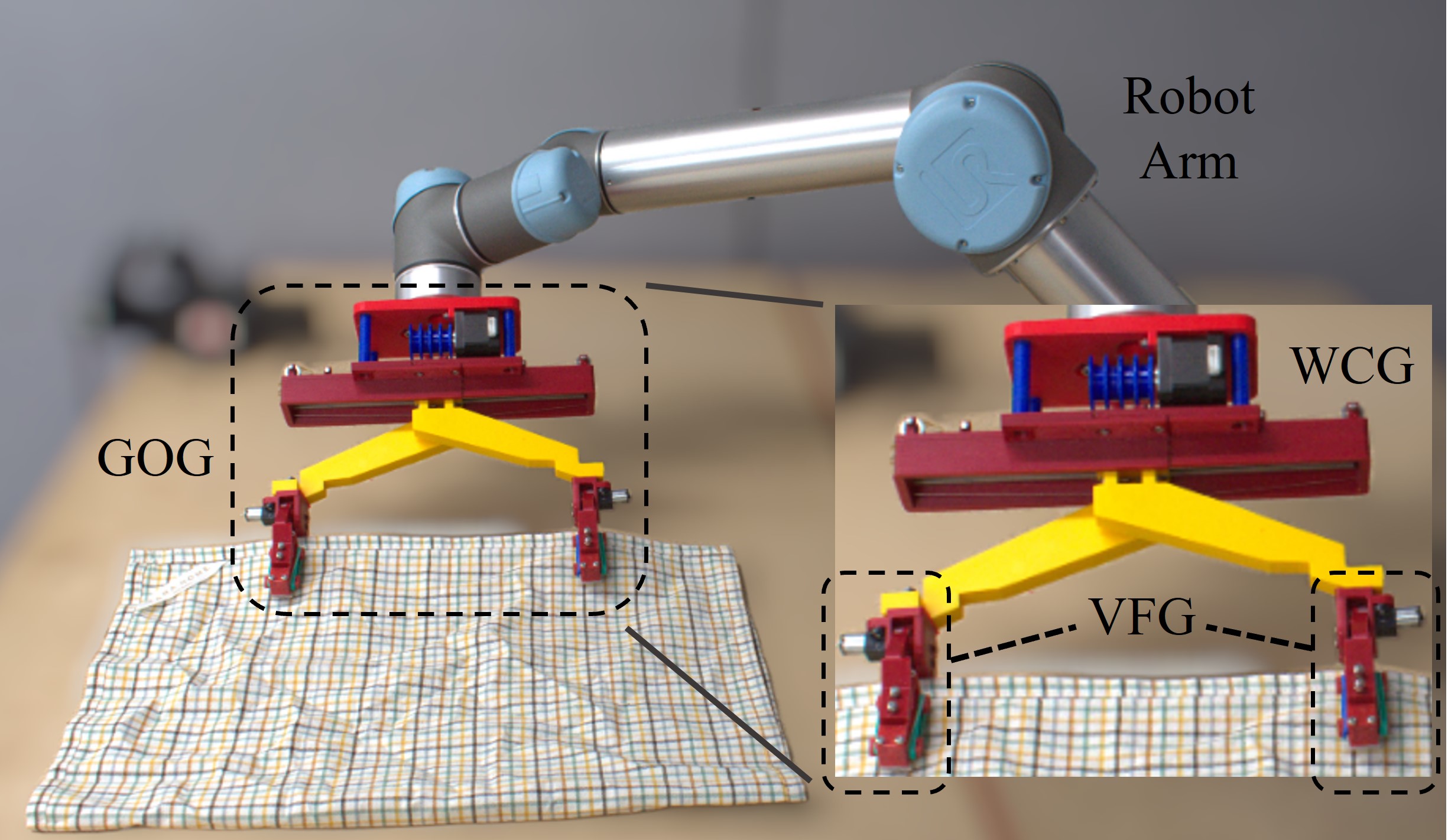}
    \caption{The {\it G.O.G.} gripper consists of a gripper base (palm), a Width Control Gripper (WCG), and two Variable Friction Grippers (VFG). The WCG can adjust the opening of fingers to fit various sizes of cloth, while the VFG mounted on these fingers can switch the friction model for the implementation of firm grasp and sliding grasp motions as necessary.}
    \label{fig:overall_system}
    \vspace{-10pt}
\end{figure}




In this work, we introduce a novel gripper design that facilitates the execution of bimanual cloth manipulation tasks, encompassing actions like folding and flattening, with the use of a single robot arm (See Fig.~\ref{fig:overall_system}). Our proposed gripper configuration comprises two distinct features: a Width Control Gripper (WCG) and two Variable Friction Grippers (VFG). The overall gripper follows then a Gripper-on-Gripper ({\it G.O.G.}) structure. The WCG functions as a large-scale gripper mechanism responsible for regulating the width of the two VFGs. This design enables the gripper to grasp two corner regions of the intended fabric, accommodating a wide range of sizes from small to large. In the context of the VFG, it functions in two distinctive modes: grasping mode and sliding mode. This is achieved via the integration of a variable friction (VF) module~\cite{spiers2019using,8989813}, which enables both power grasping and sliding grasping of the cloth. The power grasping mode ensures a safe grasp by means of high friction interaction, while the sliding mode enables the cloth to be flattened using a low friction contact. The transition between these modes is activated passively, depending on the applied gripping force.

The evaluation of the proposed approach centers on two aspects: (i) a performance assessment of the {\it G.O.G.} gripper itself, encompassing payload capacity, lifting capabilities, and accuracy in placing objects, and (ii) a comprehensive evaluation of the robotic system as a whole, specifically in terms of folding accuracy of benchmark household fabrics. The experiment results not only demonstrate gripper performance on par with reference commercial grippers but also show the capabilities of the gripper to proficiently execute various cloth manipulation tasks, such as folding, flattening, and hanging. The rest of the paper unfolds as follows: The upcoming section introduces the state of the art end-effector designs and dual-arm robotic systems associated with the cloth manipulation tasks. Section III provides an in-depth look at the design details of the proposed gripper. Meanwhile, Section IV demonstrates the capabilities of our proposed gripper design. Finally, Sections V and VI encompass the discussion and conclusion, respectively.




%
\section{Related works}
\subsection{End-effector Design for Cloth Manipulation}
The manipulation of garments poses a significant challenge in the field of robotics due to their pronounced deformability. Nevertheless, a notable proportion of cloth manipulation grippers are primarily designed to achieve reliable grasping rather than facilitating intricate manipulation tasks~\cite{borras2020grasping}. Grippers designed for cloth manipulation, which incorporate features like needles~\cite{kondratas2005robotic}, rollers~\cite{sahari2010clothes}, and suction cups~\cite{cubric2019study}, have been devised to simplify the process of gripping distinct layers of fabric and identified corners. However, these specialized grippers can be challenging to adapt for broader manipulation tasks beyond their designated functions.
To tackle this challenge, parallel jaw grippers~\cite{thuy2013development, dragusanu2022dressgripper} and multi-fingered hands~\cite{ono2007better, koustoumpardis2014underactuated, donaire2020versatile} have been put forth to enable broader cloth manipulation tasks, including folding, pick-and-place, unfolding or flattening. However, the majority of these gripper designs are not inherently designed to manipulate the cloth using only a single robot arm. 
In this letter, we present a gripper specifically designed for bimanual manipulation, especially oriented towards tasks like folding and flattening. This gripper is intended to be used with a single robot arm and incorporates a variable friction module, enabling the transition between secure grasping and sliding operations.

Our gripper shares similarities with~\cite{shibata2008handling} as a linear-actuated gripper, featuring two compact parallel grippers and the capacity to execute a pinch-and-slide motion for the tasks involving flattening. Nevertheless, the gripper in~\cite{shibata2008handling} is characterized by a substantial weight resulting from the incorporation of ball screws and linear guides. Furthermore, it necessitates additional movements to intentionally create wrinkles into the fabric for successful grasping on a table. This entails actions like applying pressure to the fabric edge with two fingertips and exerting horizontal pressure on the surface using two grippers to create wrinkles.
In contrast, our gripper entails a novel linear gripper design that leverages (i) a tendon-driven mechanism and 3D-printed components to reduce both the dimensions and weight of the gripper and (ii) a sliding component located on each fingertip, facilitating the secure grasping of fabric on a table without requiring additional maneuvers.

\subsection{Cloth Manipulation with Dual-arm Robotic System}
Several approaches have been pursued to address cloth folding tasks, including fold on a table, fold in the air, and dynamic fold in the air. In the context of this research, our focal manipulation objective is fold on the table, encompassing the process of grasping two vertices and folding the cloth until the desired fold configuration is achieved~\cite{miller2012geometric, li2015folding, doumanoglou2016folding}. Planning such folds necessitates not only the recognition of the cloth itself and determining the optimal placement of folding lines~\cite{miller2012geometric}, but also the computation of gripper trajectories required to execute the manipulation~\cite{li2015folding}. Yet another challenge arises in the need to monitor the folding process of the cloth in a controlled manner. This could involve employing visual feedback\cite{jia2018manipulating} or implementing force control manipulation to maintain tension between hands~\cite{lee2015learning} during the folding procedure. However, these methodologies involve bimanual manipulation with the use of a pair of robot arms, introducing complexity and difficulty in control.
In~\cite{sugiura2010foldy}, the authors endeavored to address this manipulation challenge using a single linear gripper. Nevertheless, this approach is unsuitable for folding cloths of varying sizes, as it presents a considerable challenge to ascertain whether the corners of the target garment have been securely grasped or not. This uncertainty leads to unpredictable folding outcomes.


\begin{figure*}[t!]
    \centering
    \vspace{4mm}
    \includegraphics[width=0.9\textwidth]{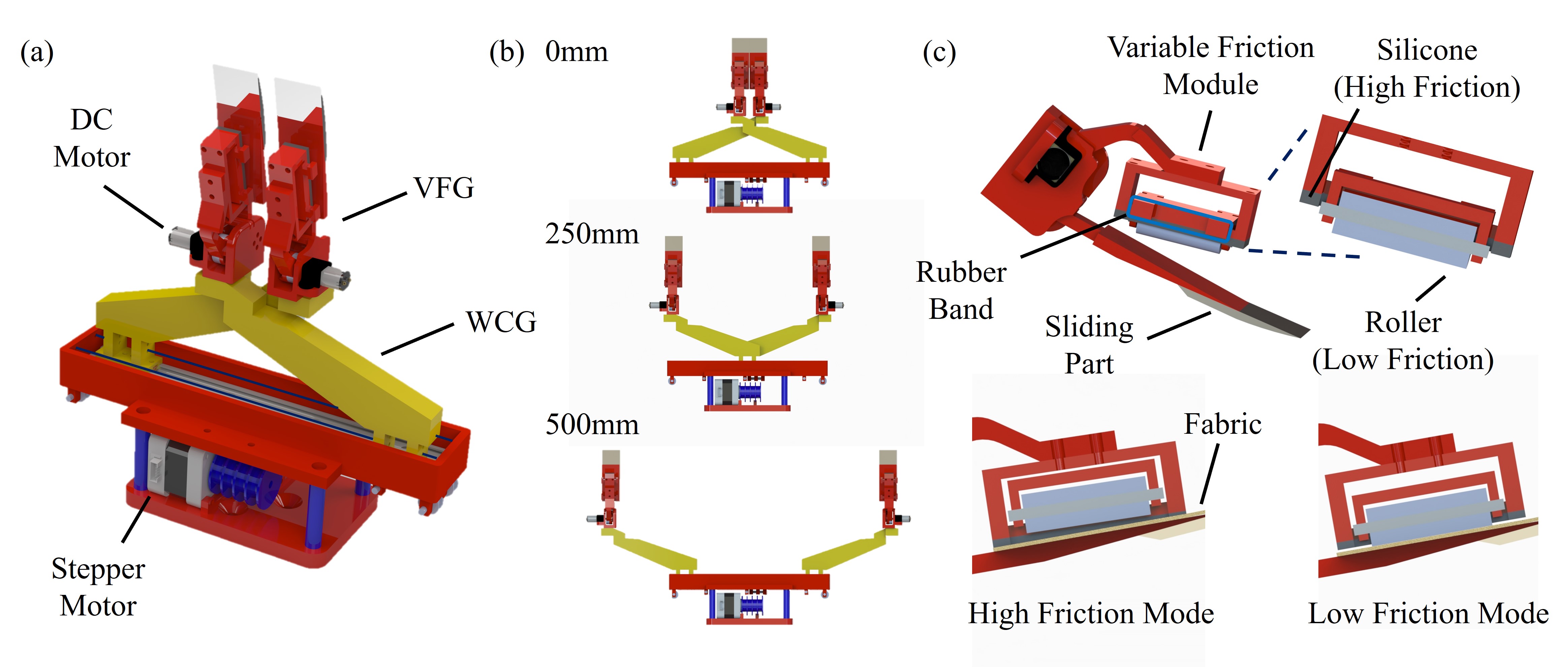}
    \caption{The {\it G.O.G.} gripper overview: \textbf{(a)} the CAD model of the {\it G.O.G.} gripper, showing the base, width control gripper (WCG), and variable friction grippers (VFG), \textbf{(b)} the range of WCG opening width, and \textbf{(c)} VFG gripper, detailing the variable friction module (passively switch between high friction and low friction modes), actuation, and sliding part.}
    \label{fig:GOG_overall_figure}
    \vspace{-10pt}
\end{figure*}

Cloth flattening on a table has predominantly centered on the domains of perception and control. For instance, several approaches involve sliding the gripper surfaces on top of the cloth~\cite{cuen2008action, maitin2010cloth, doumanoglou2016folding}, while others entail grasping its edges or vertices and subsequently applying a pulling force~\cite{willimon2011model, sun2015accurate}. These studies have addressed to localize the state of the garment and identify creases, thus facilitating the determination of appropriate directions for sliding or pulling actions that contribute to the flattening of the cloth. In particular, in~\cite{doumanoglou2016folding}, these studies demonstrated the efficacy of a special-designed gripper tailored for cloth manipulation~\cite{thuy2013development} in successfully accomplishing cloth flattening tasks, but it is still challenging to achieve the bimanual manipulation using just a single robot arm.
In terms of gripper design, a notable approach involves the integration of a variable friction module~\cite{donaire2020versatile}. This module is applied to facilitate smoother sliding motions, thereby enhancing the gripper's ability to execute its intended tasks. They employed a cam mechanism, utilizing a small motor, to trigger the high friction mode. This distinctive configuration indicates that the actuation of grasping and the variable friction module is governed by a separate motor.
In general, the majority of research endeavors have concentrated on devising approaches related to garment folding and flattening in terms of perception and control. While some authors have proposed novel gripper designs, these designs may be constrained in their applicability to single-armed bimanual manipulation scenarios. In our approach, we develop a novel gripper that serves a dual purpose. This gripper is not only adept at executing garment manipulation tasks using a single robot arm, capitalizing on its extensive linear motion capabilities, but it also introduces the capability for sliding motion via the implementation of a variable friction module located on each fingertip~\cite{spiers2019using, donaire2020versatile}.




\section{Methodology}
\subsection{The Gripper-on-Gripper System Overview}
In this section, we introduce the design details of the Gripper-on-Gripper ({\it G.O.G.}), as depicted in Fig.~\ref{fig:GOG_overall_figure}(a). The {\it G.O.G.} is composed of two distinct modules: (i) the Width Control Gripper (WCG), responsible for regulating the width of two compact end-effectors, and (ii) the Variable Friction Gripper (VFG), designed to seamlessly transition the gripper's operational modes between a secure grasp and a sliding motion. Two VFGs consistently operate in identical friction modes: utilizing the low friction mode for the flattening task and the high friction mode for the folding and hanging tasks, respectively.

\subsubsection{\textbf{Width Control Gripper}}
To fit the various dimensions of the target cloth, we introduce a Width Control Gripper (WCG) module to adjust the opening width of two end-effectors, as shown in Fig.~\ref{fig:GOG_overall_figure}(b).
The WCG is actuated by a single stepper motor (NEMA 17) using tendons, which serves to reduce the dimensions of the {\it G.O.G.} and simplify the control strategy. The transmission system efficiently transmits the motor's force to each individual finger of the WCG via the utilization of tendons connecting the motor to the fingers. A linear guide is employed to ensure the smoothness of each finger's movement, thereby minimizing friction during both forward and backward motions. The tendon-driven mechanism offers the advantage of a wide range of finger width adjustments without adding bulkiness to the gripper. To optimize the motion range of each finger, a design has been implemented that ensures the avoidance of collisions between two fingers. According to the benchmark from~\cite{garcia2022household}, we take the maximum and minimum values of the cloth size for our design choices.
Taking into account that each finger can move within a range of 300mm, this design not only allows two VFGs to function as a single gripper by contacting each other, but it also enables them to extend outward to approximately 500mm. This extension capability facilitates bimanual manipulation using a single robot arm.



\subsubsection{\textbf{Variable Friction Gripper}}
For most rigid-body manipulation, a simple pinch grasp can implement most of the manipulation tasks. However, manipulating cloth, such as cloth edge-tracing and in-air flattening, requires a more complex pinch-slide motion for its execution. 
In this work, the VFG is proposed to achieve such secure grasping and flattening motion passively by controlling the torque of DC motors, as illustrated in Fig.~\ref{fig:GOG_overall_figure}(c). In the context of the variable friction module, a roller is designed to serve as a low-friction component that aids in the flattening task while the WCG is in motion. As the torque of the DC motor increases, the mode of the VFG changes to high friction, signifying that the gripper is capable of firmly attaching to a garment using a silicone part, thus achieving a secure grasp. This passive transition is enabled by a rubber band that functions like a spring, pushing the roller outward when the motor torque increases. Otherwise, it maintains a low friction mode, with the roller maintaining contact with the fabric. Our variable friction mechanism is based on~\cite{8411094}; however, our design does not require any actuation for the friction switching, which can simplify the gripper control and circuit design complexity.

Garments always employ a two-layer or multi-layer structure. In this case, a top-down pinch grasp will easily cause failure by grasping undesired layers of cloth. Therefore, in our work, we design a sliding grasping module for cloth grasping. 
The sliding component of the VFG is intended for grasping fabric by smoothly sliding it beneath the fabric. To achieve this, a TPU material is employed due to its combination of low friction and high elasticity. The choice of a DC motor is driven by its compact size and robust torque characteristics, enabling it to effectively grasp even heavy garments such as bedsheets and blue jeans.





\subsection{{\it G.O.G.} for Bimanual Cloth Manipulation Tasks}
\subsubsection{\textbf{Automatic Bimanual Cloth Folding}}
\begin{figure}[t!]
    \centering
      \vspace{4mm}
    \includegraphics[width=0.95\columnwidth]{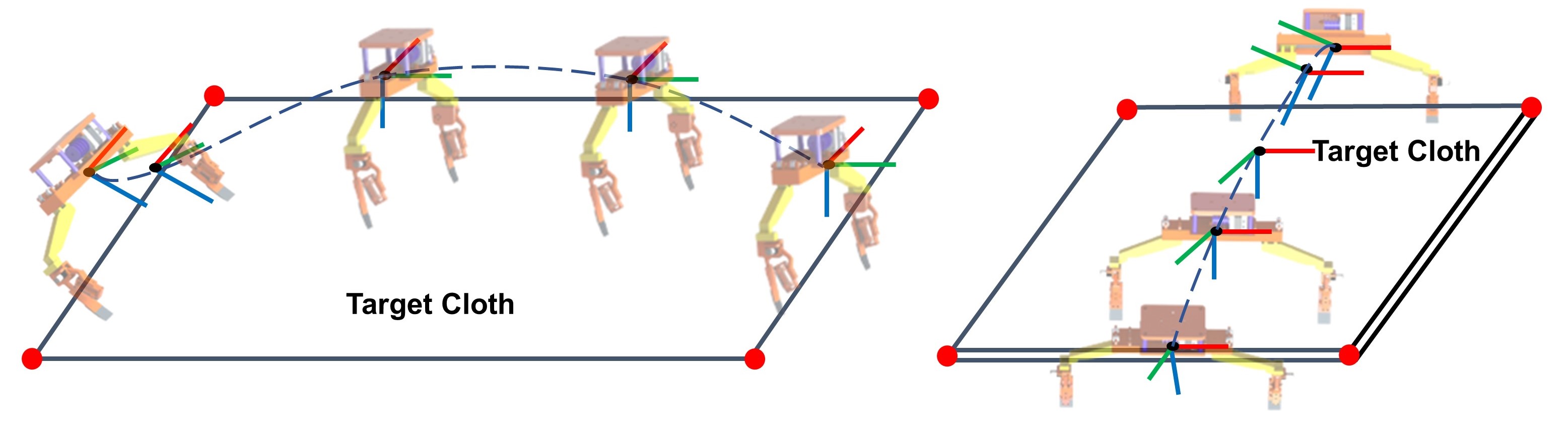}
    \caption{Folding trajectories based on the detected corner point. The Width Control Gripper (WCG) adjusts the opening width firstly. Then, pre-grasp, grasp, folding, and release actions are executed. The VFGs are switched to high friction to ensure a firm grasp during the folding.}
    \label{fig:Figure/Figure_folding_principle.jpg}
    \vspace{-10pt}
\end{figure}

One major advantage of our design is the implementation of bimanual cloth manipulation by employing just one robot arm. Our proposed gripper design can significantly simplify the complex motion planning procedure for cloth manipulation. In this paper, we use the bimanual cloth folding tasks as the case study to demonstrate the effectiveness of our gripper design in facilitating cloth manipulation tasks. Unlike previous research that requires coordination of two robot arms~\cite{weng2022fabricflownet}, we can implement the cloth folding control by simply adjusting the gripper pose and its opening width.
Our overall folding system is depicted in Fig.~\ref{fig:Figure/Figure_folding_principle.jpg}. To simplify the overall control process, we implement the perception system with a classical corner detection algorithm. With an overhead Realsense RGB-D camera, we can locate the corner points of the target cloth for the cloth grasping. We then define a set of folding trajectories based on the detected corner points. In the meantime, the grasp width is also adjusted according to the distance of two corner points. Our design is to ensure a proper distance between two small VFGs to limit the corner deformation for better folding quality. An evaluation of the bimanual cloth folding system is conducted in the following sections.

\subsubsection{\textbf{Bimanual Cloth Hanging and Cloth Edging Flattening}} 

\begin{figure}[t!]
\centering
\vspace{4mm}
\subfigure[Cloth Hanging]{
\begin{minipage}[t]{0.95\linewidth}
\centering
\includegraphics[width=1\columnwidth]{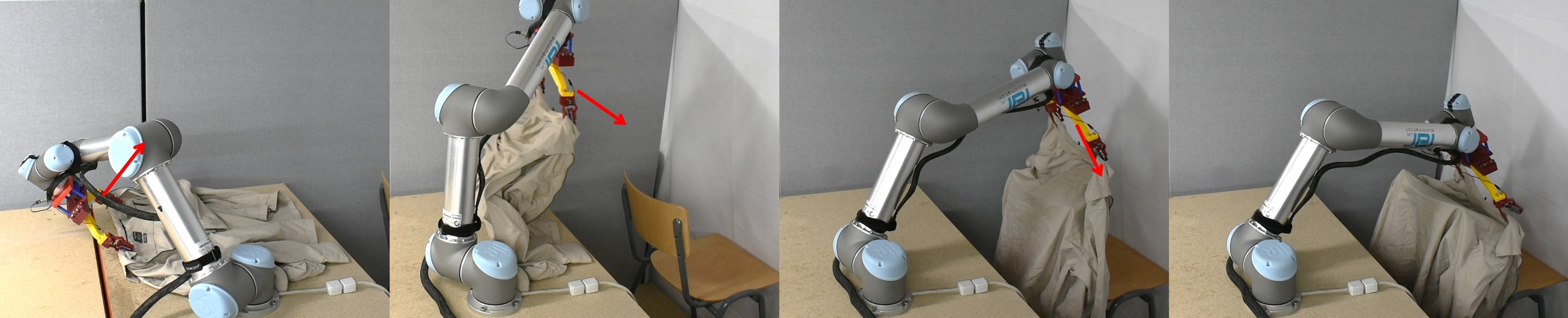}
\end{minipage}}
\centering
\subfigure[Cloth Flattening]{
\begin{minipage}[t]{0.95\linewidth}
\centering
\includegraphics[width=1\columnwidth]{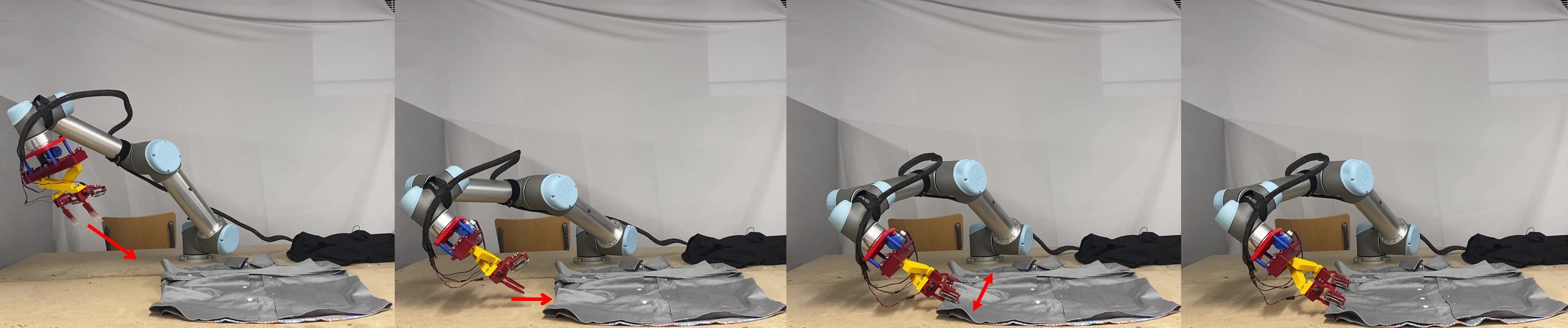}
\end{minipage}%
}

\centering
\caption{(a) Example of a hanging task. The task is to grasp the collar of the garment and hang it on the back of the chair.
(b) Example of cloth flattening. The task is to grasp the wrinkle area of the garment and slide over the wrinkle to make the garment flattened.}
 \label{fig:flatten and hanging}
 \vspace{-10pt}
\end{figure}

We further explore the potential of our {\it G.O.G.} gripper by doing a bimanual cloth hanging task and cloth flattening. For the garment hanging task, a fixed configuration of the garment is applied. The following pre-defined motion, including grasp, lift, transport and release, is executed to finish the task. 

We then performed a cloth flattening task with the {\it G.O.G.} gripper. A similar fixed configuration of the garment is also applied in this task. In the firm grasp stage, we employed a high-friction silicone pad to ensure a robust grasping force. Subsequently, for the flattening motion, we applied a reduced torque for the sliding grasp. The roller then took on the primary role as the main point of contact to facilitate smooth sliding motion.

\section{Experiments}
To demonstrate the capabilities of our proposed gripper design, we conduct an evaluation of the {\it G.O.G.} using two fabric benchmarks: household items~\cite{garcia2022household} and clothing objects~\cite{clark2023household}. To be more precise, our assessment comprised three individual experiments: (1) The payload capacity of the {\it G.O.G.} design, (2) Cloth grasp-and-lift experiment, (3) Cloth drag placement accuracy, and (4) Cloth manipulation tasks to effectively demonstrate the prowess of our gripper design and its integration with a robotic system. This comprehensive evaluation serves to highlight the capabilities and performance of our innovative gripper design.

\begin{figure}[t!]
    \centering
    \vspace{4mm}
    \includegraphics[width=0.95\columnwidth]{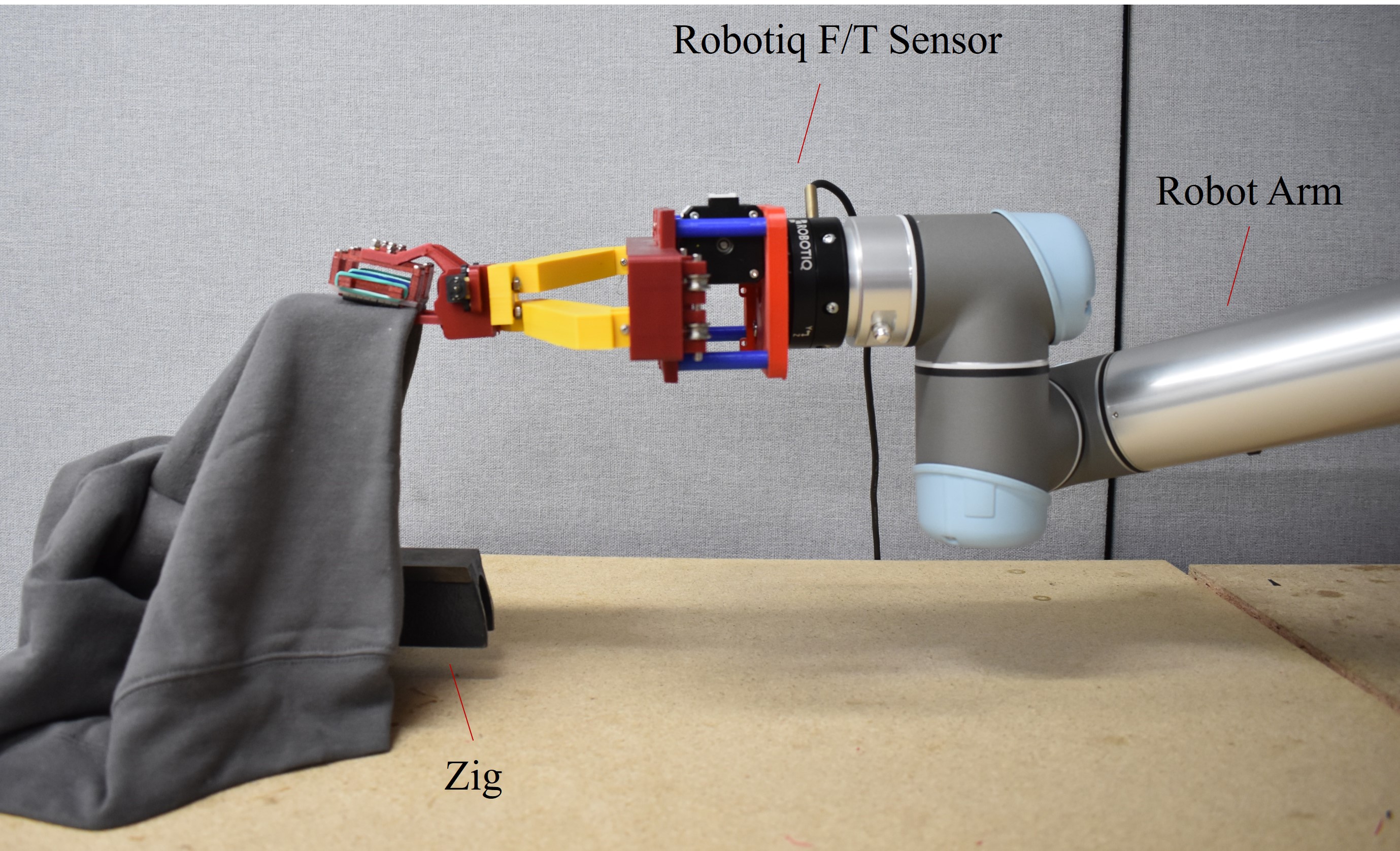}
    \caption{The payload experimental setup: UR5 and Robotiq FT 300 sensor are employed.}
    \label{fig:Payload_setup}
    \vspace{-10pt}
\end{figure}

\subsection{Gripper Payload Capacity}
\begin{figure}[t!]
    \centering
    \vspace{4mm}
    \includegraphics[width=0.98\columnwidth]{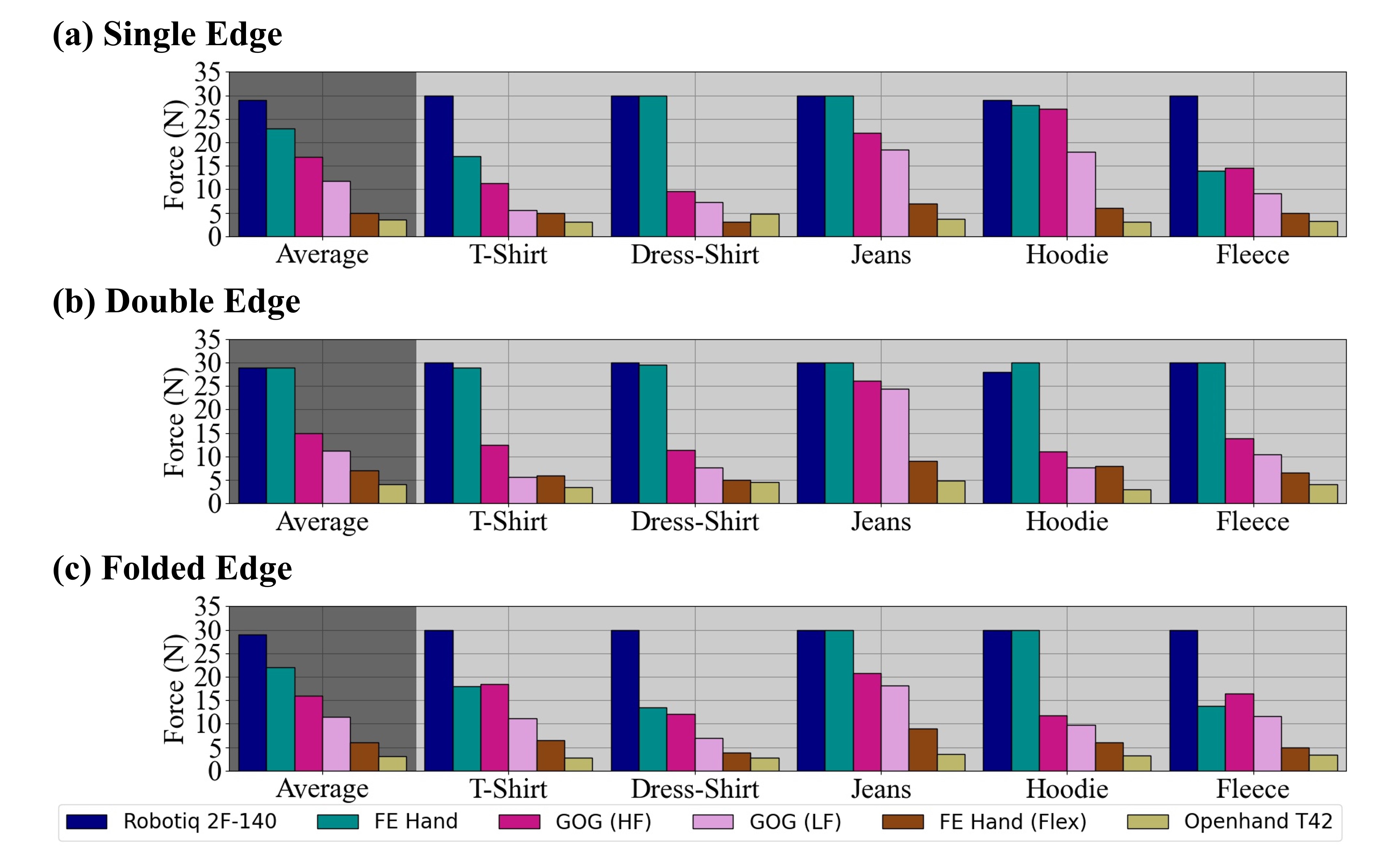}
    \caption{Payload result of the {\it G.O.G.} in terms of high friction module and low friction module. Higher force means better payload performance.}
    \label{fig:Payload_result}
    \vspace{-6mm}
\end{figure}

We evaluate the payload capability of the gripper using a Force/Torque sensor (Robotiq FT 300). The experimental setup is designed to ensure a fair comparison, similar to the setup used in~\cite{clark2023household}. As in Fig.~\ref{fig:Payload_setup}, both the FT sensor and the gripper are attached to the UR5 robot to move linearly away, while a piece of fabric is securely fixed to a rigid platform. The payload capacities of the high friction module (HF module) and the low friction module (LF module) are depicted in Fig.~\ref{fig:Payload_result}. We set the maximum pulling force at 30N, which is chosen to not only exceed the weight of any cloth but also to prevent potential damage to the items being manipulated. The resulting force is obtained by capturing the peak force before a grasp fails. For those grasps that do not fail, a 30N will be recorded. As suggested in~\cite{clark2023household}, only objects with a minimum mass of 0.1 kg will be selected.


\subsection{Cloth Grasp-and-Lift}
\begin{figure}[t!]
    \centering
      \vspace{4mm}
    \includegraphics[width=0.98\columnwidth]{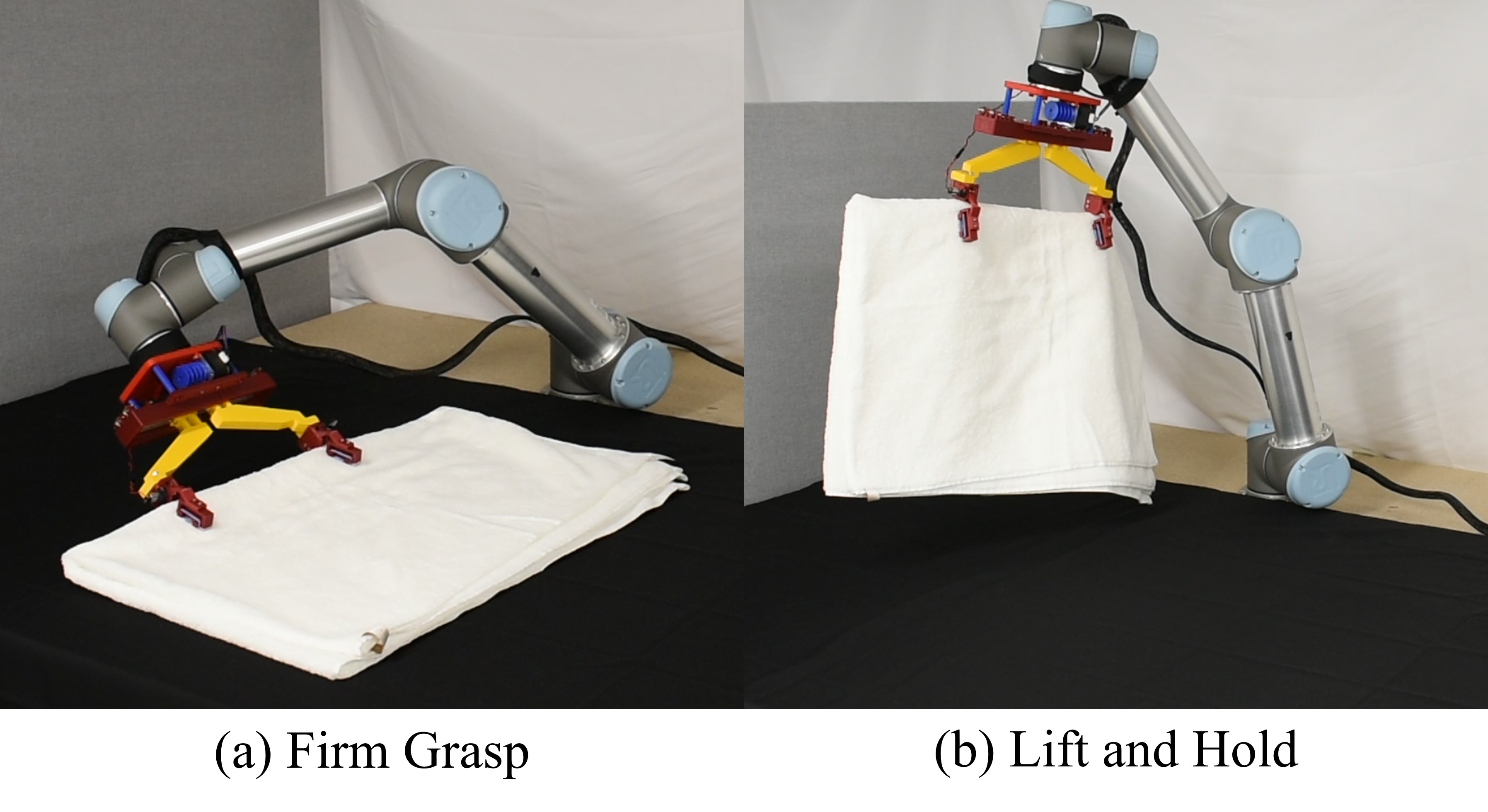}
    \caption{Grasping experiment setup of the {\it G.O.G.} in terms of high friction module.}
    \label{fig:Grasp_setup}
    \vspace{-6mm}
\end{figure}

Having already completed the payload experiment with the cloth object benchmark, we utilized the household item benchmark to assess the grasp-and-lift capacity. This assessment allows us to comprehensively evaluate the gripper's performance in terms of varying sizes and material properties. An overview of this benchmark can be found in~\cite{garcia2022household}. As depicted in Fig.~\ref{fig:Grasp_setup}, the initial state of the target fabric involves it lying flat on a table without any creases. To determine the success of cloth grasp-and-lift, we lifted the fabric off the table surface to a height of 500mm after securely grasping it, followed by a holding motion in this position for a duration of 5 seconds. The results of the grasp-and-lift success rate are presented in Table. 1.
\par
The classification of grasp-and-lift success is divided into three categories: perfect, half, and fail. A perfect grasp denotes that both VFGs have successfully grasped and lifted the fabric object securely, a half grasp indicates that only one VFG has managed to secure the grasp-and-lift motion, and a fail implies that this motion is not accomplished successfully. 

\begin{figure}[t!]
\centering
\vspace{4mm}
\includegraphics[width=0.98\columnwidth]{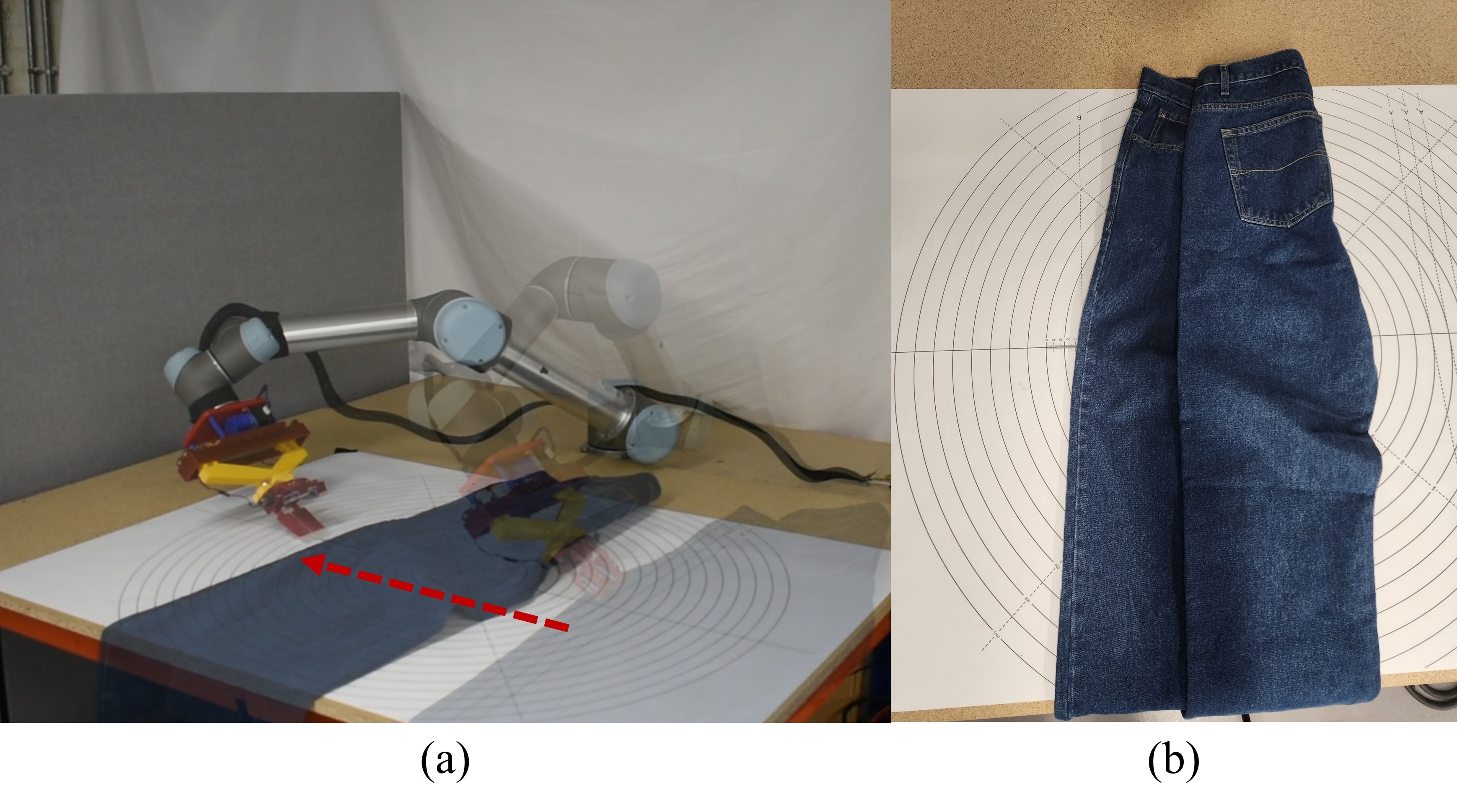}
\caption{Drag experiment setup of the {\it G.O.G.} (a) The robot execution of dragging the folded edge of jeans. (b) The result is obtained by an alignment grid}
 \label{fig:drag_demo}
 \vspace{-6mm}
\end{figure}

\begin{table}[b!]
    \vspace{-4mm}
    \centering
    \caption{Grasp-and-Lift success rate for the household benchmark (10 trials for each object).}
    \resizebox{0.8\columnwidth}{!}{%
    \begin{tabularx}{\columnwidth}{| l | l | >{\centering\arraybackslash}X | >{\centering\arraybackslash}X | >{\centering\arraybackslash}X |}
        \hline
        \multirow{2}{*}{\bf Category} & \multirow{2}{*}{\bf Name} & \multicolumn{3}{c|}{\bf Grasp-and-Lift Success (\%)} \\ [0.2ex]
        \cline{3-5}
        & &Perfect &Half &Fail\\
        \hline
        \multirow{3}{*}{Bathroom} &Small towel &100 &0 &0 \\
        \cline{2-5}
        &Med. towel &100 &0 &0 \\
        \cline{2-5}
        &Big towel &100 &0 &0 \\
        \hline
        \multirow{4}{*}{Bedroom} &Bedsheet &70 &30 &0 \\
        \cline{2-5}
        &Fitted bedsheet &70 &30 &0 \\
        \cline{2-5}
        &Sq. pillowcase &100 &0 &0 \\
        \cline{2-5}
        &Rect. pillowcase &100 &0 &0 \\
        \hline
        \multirow{4}{*}{Dining room} &Rect. tablecloth &100 &0 &0 \\
        \cline{2-5}
        &Round tablecloth &90 &0 &10 \\
        \cline{2-5}
        &Cotton napkin &100 &0 &0 \\
        \cline{2-5}
        &Linen napkin &80 &20 &0 \\
        \hline
        \multirow{4}{*}{Kitchen} &Towel rag &80 &10 &10\\
        \cline{2-5}
        &Linen rag &90 &10 &0\\
        \cline{2-5}
        &Waffle rag &100 &0 &0\\
        \cline{2-5}
        &Checkered rag &100 &0 &0\\
        \hline
    \end{tabularx}
    \vspace{-15pt}
    }
    \label{table:2}
\end{table}

\subsection{Cloth Dragging}
Dragging placement accuracy is quantified for the cloth object benchmark, enabling a comparison of performance against that of commercially available grippers. To ensure a valid comparison, our experimental setup closely adheres to the methodology in~\cite{clark2023household}. This experiment encompasses the use of an alignment grid and adherence to the same procedural steps as outlined in their study. In our evaluation, we compared the dragging placement accuracy with the established baseline. The assessment procedure unfolds as follows: (1) Begin by grasping the flat edge of each cloth object benchmark, (2) Lift the object off the surface once it is securely grasped, (3) Execute a horizontal translation of 500mm using the manipulator to drag the object, and (4) Complete the process by releasing the object. To limit the deformation of the cloth center region, we minimize the gripper opening width during the entire dragging execution. Unlike the top-down pinch grasp implemented by other off-the-shelf grippers, {\it G.O.G.} is primarily designed for a sliding grasp. This grasping strategy is also implemented in the grasp stage of this drag experiment.

As illustrated in Fig.~\ref{fig:drag_demo}, the measurement of dragging placement is conducted from the alignment line, using a 1mm increment ruler, to determine the precise location of the instantaneous point along the edge of the target garment. Following the benchmark from~\cite{clark2023household}, only items of substantial sizes that require folding execution are selected as the experimental targets.
A perfect score is attained when the edge is successfully grasped, remains held throughout the dragging motion, and is placed precisely on the alignment line (resulting in a 0mm offset). 

\begin{figure}[t!]
  \vspace{4mm}
    \centering
    \includegraphics[width=0.9\columnwidth]{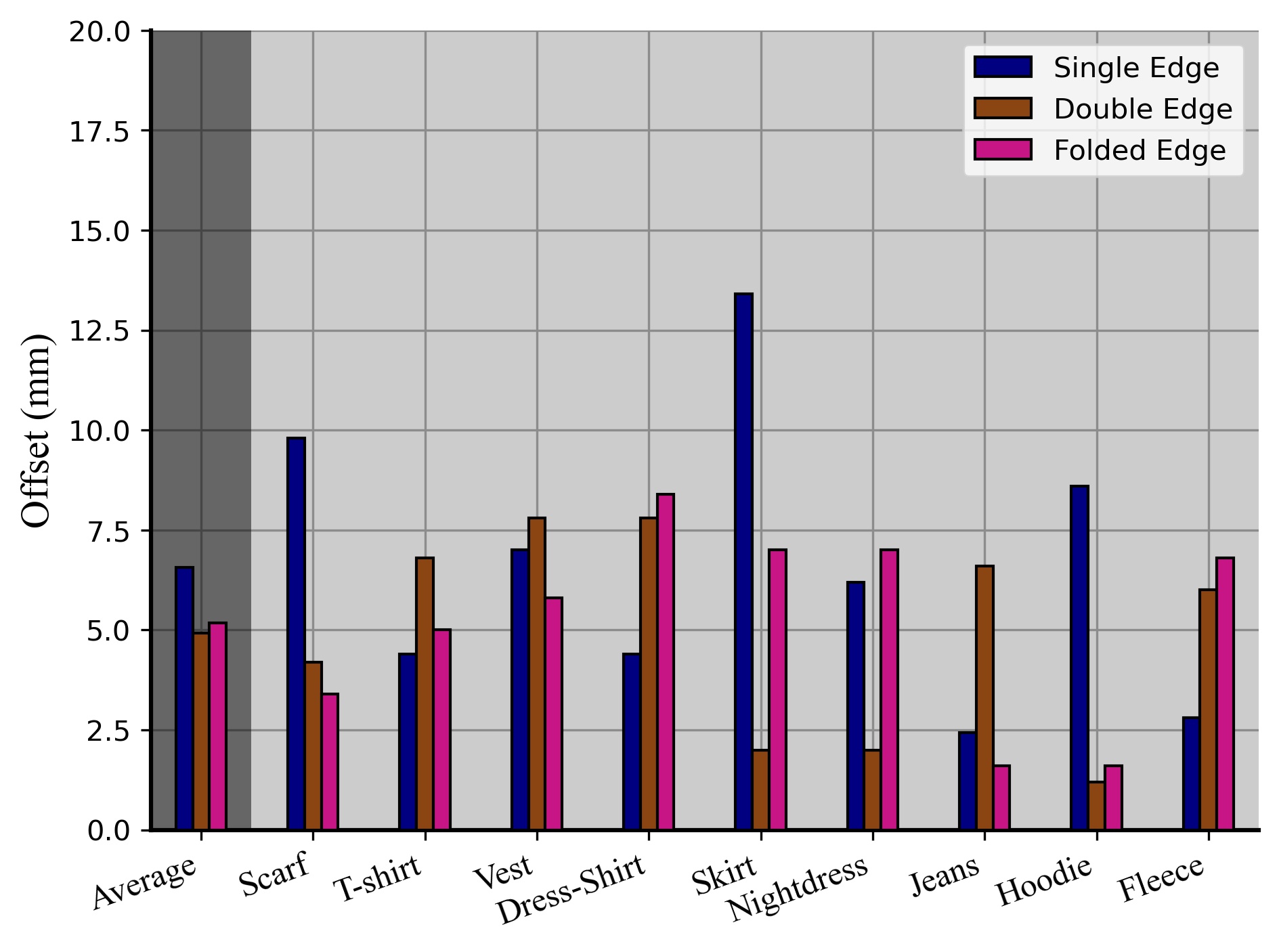}
    \caption{Drag placement accuracy of the {\it G.O.G.} The average value is shown here. Lower offset means better drag placement performance.}
    \label{fig:Fig_drag}
    \vspace{-10pt}
\end{figure}

\begin{figure}[t!]
    \centering
      \vspace{4mm}
    \includegraphics[width=0.9\columnwidth]{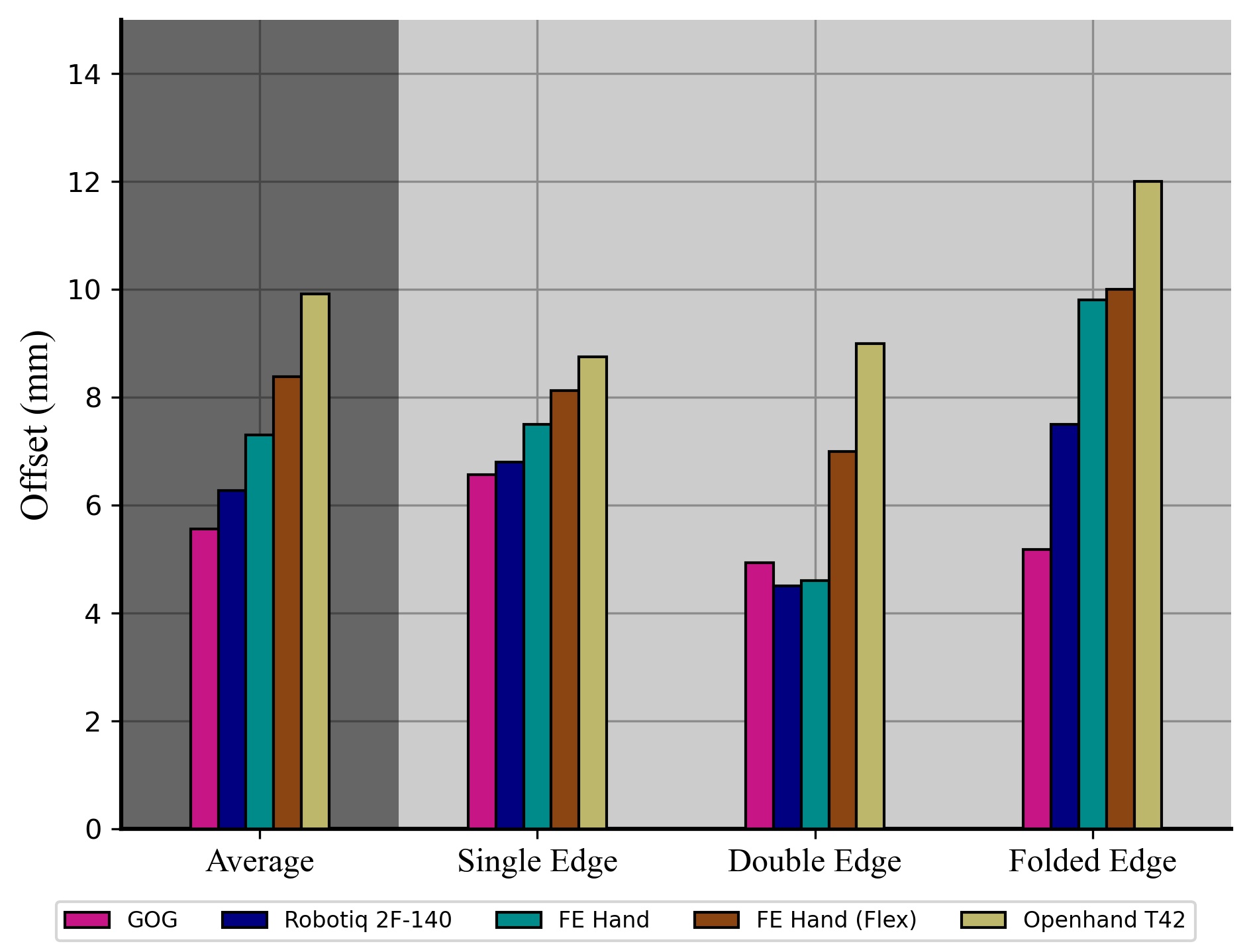}
    \caption{Average drag accuracy comparison result with off-the-shelf grippers. Single Edge, Double Edge and Folded Edge refer to different edge types for grasping~\cite{clark2023household}.}
    \label{fig:Fig_drag_comp}
    \vspace{-15pt}
\end{figure}

\subsection{Robotic System Experiment}

\begin{figure*}[t!]
    \centering
    \vspace{4mm}
    \includegraphics[width=0.98\textwidth]{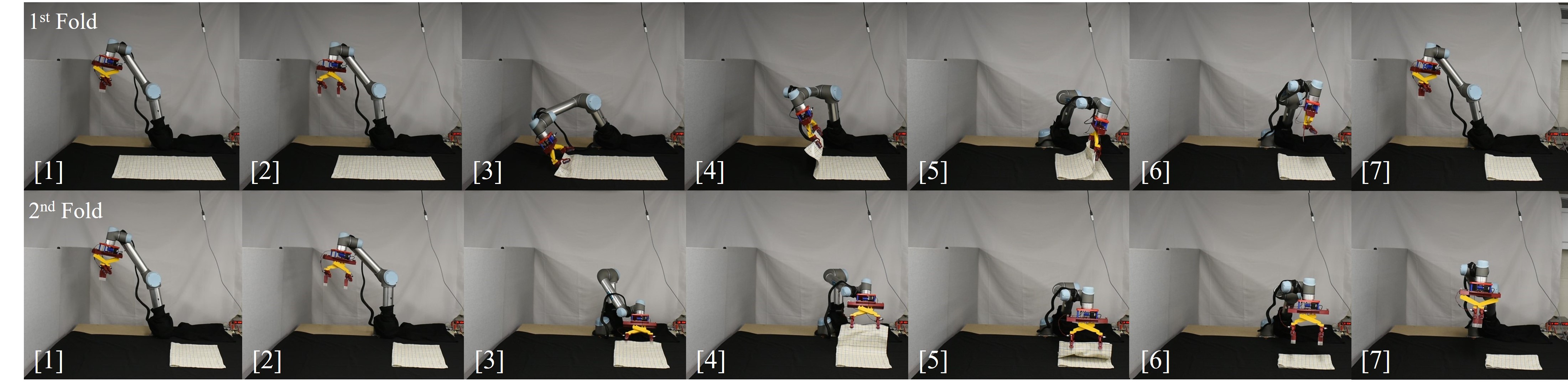}
    \caption{The real-world robot cloth folding experiment demonstration. Two columns represent the first and second robot folding in a time-series, respectively. Before each operation, the {\it G.O.G.} will initialize the opening width to the minimum position. This aims to reduce the dimension of the gripper itself to ensure enough workspace for the motion planning of the whole robot arm. Before reaching the target pose, the {\it G.O.G.} will adjust its opening width based on the cloth size by detecting two corner points with an overhead camera. The communication between devices(UR5, gripper, PC and camera) is implemented by ROS. The motion planning of the robot arm is achieved by Moveit!~\cite{chitta2012moveit}.}
    \label{fig:folding figure}
    \vspace{-12pt}
\end{figure*}

In this section, we demonstrate the performance of cloth folding to show the capability of the bimanual robotic system with our {\it G.O.G.} gripper and a single UR5 robot arm. We evaluate the bimanual cloth folding as our case study in this section. As shown in Fig.~\ref{fig:folding figure}, we follow the benchmark from~\cite{garcia2022household} to conduct the folding experiment.
Specifically, the experiment consists of two phases: the first fold (1-fold) and the second fold (2-fold). In terms of the first folding, the initial state of the target fabric is flat to measure the folding performance accurately. In addition, we also rotate the cloth into different angles to get a more robust result. Five trials, including both 1-fold and 2-fold, are performed for each object in this experiment.

Following~\cite{weng2022fabricflownet}~and~\cite{hoque2022learning}, we measure the folding accuracy by two metrics: (1) intersection over union (IoU) and (2) a penalty for edges and wrinkles (WR). Being different from other learning-based approaches, the ground truth image and its corresponding mask are not available for our method to calculate the MIoU. Therefore, we generate the desired 'ground truth' mask by halving the image's contours along the folding direction to make the measurement. After the generating procedure, IoU is then computed between the actual cloth mask and the desired 'ground truth' mask. The wrinkle penalty (WR) is achieved by detecting the wrinkle pixels within the cloth mask area. In our research, we use the Canny Edge detector for the detection~\cite{canny1986computational}. With such a definition, a high-quality folding execution needs to achieve a high IoU with a low WR for higher cloth coverage and low wrinkles inside. Five trials are performed for each target cloth to get the mean value for the final results. In addition, we also changed the position and orientation of the cloth to test our overall system in different configurations during the experiment.

\begin{figure*}[t!]
    \centering
    \vspace{4mm}
    \includegraphics[width=0.98\textwidth]{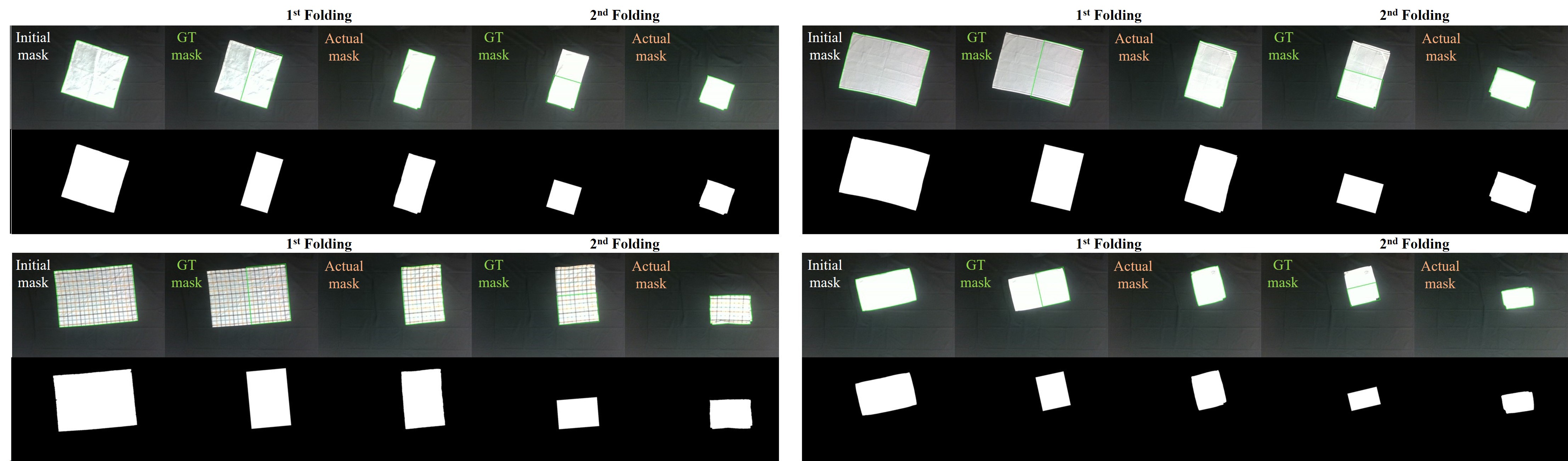}
    \caption{The real-world robot cloth folding experiment result. We compute the IoU between the generated 'ground truth' cloth mask and the actual cloth mask after each folding execution. A wrinkle penalty is also computed with the detected cloth masks.}
    \label{fig:folding image figure}
    \vspace{-10pt}
\end{figure*}

\begin{table}[t]
   \vspace{4mm}
\centering

\caption{Robotic folding evaluation with robot and {\it G.O.G.} gripper.}
\resizebox{0.9\columnwidth}{!}{%
\setlength{\tabcolsep}{3mm}{
\begin{tabular}{c| c c }
\hline
\textbf{Name}  &\textbf{1-fold (MIoU / WR)} & \textbf{2-fold (MIoU / WR)}   \\ [0.5ex]
\hline
\makecell[c]{Waffle rag 1}       &0.939 / 0.00395 &0.929 / 0.00519\\
\hline
\makecell[c]{Checkered rag 1}     &0.834 / 0.00965 &0.906 / 0.00906\\
\hline
\makecell[c]{Checkered rag 2}         &0.926 / 0.00400 &0.892 / 0.00551 \\
\hline
\makecell[c]{Med. towel} &0.949 / 0.00352 &0.859 / 0.00484 \\
\hline
\makecell[c]{Linen rag}    &0.937 / 0.00389 &0.884 / 0.00537 \\
\hline
\makecell[c]{Waffle rag 2}    &0.936 / 0.00376 &0.896 / 0.00523 \\
\hline
\makecell[c]{Linen napkin}    &0.899 / 0.00451 &0.901 / 0.00601 \\
\hline
\makecell[c]{Sq. pillowcase}    &0.953 / 0.00340 &0.910 / 0.00428\\
\hline
\makecell[c]{Small towel}    &0.845 / 0.00530&0.640 / 0.00719 \\
\hline
\makecell[c]{\textbf{Overall}}    &\textbf{0.917} / \textbf{0.00467}&\textbf{0.868} / \textbf{0.00585} \\
\hline
\end{tabular}
}
}
\par
\begin{flushleft}
We use the Mean value of Intersection over Union (MIoU) of the result and desired image masks and WR (wrinkle penalty) to evaluate our gripper design. The value of Overall represents the average result of all the experiment trails.

\end{flushleft}
\label{table:folding result}
\vspace{-6mm}
\end{table}


\section{Discussion}
In this section, a detailed discussion is conducted for each experiment conducted in our study.

\textbf{Payload Capacity:} According to this experiment, we establish that the payload capacity of the {\it G.O.G.} is notably superior to that of customized grippers commonly used for cloth manipulation, such as Openhand T42 and Franka Emika hand with Fin Ray\textsuperscript{\textregistered} style fingers (FE hand (Flex)), as depicted in Fig.~\ref{fig:Payload_result}. This outcome confirms that the {\it G.O.G.} is well-suited for handling cloth manipulation tasks, given its remarkable payload capability.

\textbf{Grap-and-lift:} In the majority of cases, household objects can be grasped and lifted successfully, except for fabrics with larger dimensions or surfaces featuring low friction. When dealing with materials characterized by low friction or high deformability, the sliding motion underneath the cloth might not always work flawlessly, resulting in one of the VFGs fails to securely grasp the intended target object. In the case of lightweight items, a single VFG can successfully lift and maintain a hold on the entire object. However, achieving a successful lift-and-hold motion becomes notably more challenging when faced with heavier objects like bedsheets and tablecloths.

\textbf{Cloth Dragging:} The outcomes of the dragging placement accuracy assessment are illustrated in Fig.~\ref{fig:Fig_drag}. In terms of average offset, the cases of Double edge and Folded edge exhibit almost the same performance, surpassing the single edge case. This can be attributed to the fact that the overall thickness of Double Edge and Folded Edge configurations is typically similar and greater than that of the Single Edge. In addition, a comparison of the average offset results with the baseline grippers is presented in Fig.~\ref{fig:Fig_drag_comp}. In relation to Single Edge and Folded Edge scenarios, our gripper demonstrates superior performance compared to other options, except for the case of Double Edge. This is primarily due to the fact that one of the VFGs occasionally grasps only the top-side edge, leading to a less secure dragging transition.

\textbf{Bimanual Cloth Folding:} The result of the bimanual cloth folding is illustrated in Table~\ref{table:folding result}. With the {\it G.O.G.} gripper design, our robotics system can implement an overall MIoU of 0.917 and 0.868 for the 1-fold and 2-fold, respectively. Regarding the WR, our system can also reach values of 0.00467 and 0.00585 for these two stages. Several result images are illustrated in Fig.~\ref{fig:folding image figure}. The result indicates that, with the {\it G.O.G.} gripper design, the complex bimanual cloth folding tasks can be implemented with a simple corner detection algorithm and pre-defined hard-coded trajectories.   
However, a slight decrease in performance can also be observed at the 2-fold stage. This can be attributed to several factors: (1) Complexity of Multiple Folds: it is more challenging to maintain precise alignment and symmetry throughout the 2-fold stage. (2) Accumulated Errors: with the error accumulated in the 1-fold stage, the second folding is more challenging to achieve.

\section{Conclusion}

In this work, we present a novel robot gripper design named G.O.G. for bimanual cloth manipulation with one robotic arm. While employing a dual-arm system increases the overall control complexity and cost, our proposed solution allows to implement complex bimanual manipulation by just controlling gripper pose and opening width. Several experiments are conducted for the evaluation of the gripper. Specifically, we first assess the gripper's performance against several established benchmarks for cloth manipulation end-effector designs. The results indicate that the G.O.G. gripper shows superior performance in terms of cloth grasp payload, cloth grasp and lift, and cloth drag experiments against several off-the-shelf grippers. Next, we demonstrate the potential of bimanual cloth manipulation with the use of the G.O.G. gripper. Furthermore, as a case study, we evaluate the bimanual robotic cloth manipulation system in the context of cloth folding tasks. Results indicate that the challenging bimanual cloth folding task can be successfully implemented with the introduced design by only applying a simple corner detection algorithm and predefined hard-coded motion. 
While the proposed approach offers several advantages, there are limitations when it comes to folding or flattening a cloth considerably larger than the maximum width accommodated by the {\it G.O.G.}, such as a bed sheet. The proposed gripper can significantly decrease experimental costs and control complexity, however, handling larger fabrics may present challenges due to (i) the restricted dimensions of the {\it G.O.G.}, and (ii) the workspace limitations of a robot arm.
For future work, further attention could be directed towards refining the methodology from both the design and control perspectives, aiming to handle larger fabrics via minimal adjustments to the gripper dimensions.

\section{Acknowledgment}

The authors would like to thank Dr. Angus Clark for benchmark result data on the commercial gripper used in the experiment section of this work.

\addtolength{\textheight}{-2cm}   


\bibliographystyle{IEEEtran}
\bibliography{references}

\begin{thebibliography}{10}
\providecommand{\url}[1]{#1}
\csname url@rmstyle\endcsname
\providecommand{\newblock}{\relax}
\providecommand{\bibinfo}[2]{#2}
\providecommand\BIBentrySTDinterwordspacing{\spaceskip=0pt\relax}
\providecommand\BIBentryALTinterwordstretchfactor{4}
\providecommand\BIBentryALTinterwordspacing{\spaceskip=\fontdimen2\font plus
\BIBentryALTinterwordstretchfactor\fontdimen3\font minus \fontdimen4\font\relax}
\providecommand\BIBforeignlanguage[2]{{%
\expandafter\ifx\csname l@#1\endcsname\relax
\typeout{** WARNING: IEEEtran.bst: No hyphenation pattern has been}%
\typeout{** loaded for the language `#1'. Using the pattern for}%
\typeout{** the default language instead.}%
\else
\language=\csname l@#1\endcsname
\fi
#2}}

\bibitem{borras2022effective}
J.~Borr{\`a}s, ``Effective grasping enables successful robot-assisted dressing,'' \emph{Science robotics}, vol.~7, no.~65, p. eabo7229, 2022.

\bibitem{chi2021garmentnets}
C.~Chi and S.~Song, ``Garmentnets: Category-level pose estimation for garments via canonical space shape completion,'' in \emph{The IEEE International Conference on Computer Vision (ICCV)}, 2021.

\bibitem{ren2023autonomous}
Y.~Ren, R.~Chen, and Y.~Cong, ``Autonomous manipulation learning for similar deformable objects via only one demonstration,'' in \emph{Proceedings of the IEEE/CVF Conference on Computer Vision and Pattern Recognition}, 2023, pp. 17\,069--17\,078.

\bibitem{10342086}
W.~Chen, D.~Lee, D.~Chappell, and N.~Rojas, ``Learning to grasp clothing structural regions for garment manipulation tasks,'' in \emph{2023 IEEE/RSJ International Conference on Intelligent Robots and Systems (IROS)}, 2023, pp. 4889--4895.

\bibitem{sanchez2018robotic}
J.~Sanchez, J.-A. Corrales, B.-C. Bouzgarrou, and Y.~Mezouar, ``Robotic manipulation and sensing of deformable objects in domestic and industrial applications: a survey,'' \emph{The International Journal of Robotics Research}, vol.~37, no.~7, pp. 688--716, 2018.

\bibitem{sahari2010clothes}
K.~S.~M. Sahari, H.~Seki, Y.~Kamiya, and M.~Hikizu, ``Clothes manipulation by robot grippers with roller fingertips,'' \emph{Advanced Robotics}, vol.~24, no. 1-2, pp. 139--158, 2010.

\bibitem{miller2012geometric}
S.~Miller, J.~Van Den~Berg, M.~Fritz, T.~Darrell, K.~Goldberg, and P.~Abbeel, ``A geometric approach to robotic laundry folding,'' \emph{The International Journal of Robotics Research}, vol.~31, no.~2, pp. 249--267, 2012.

\bibitem{doumanoglou2016folding}
A.~Doumanoglou, J.~Stria, G.~Peleka, I.~Mariolis, V.~Petrik, A.~Kargakos, L.~Wagner, V.~Hlav{\'a}{\v{c}}, T.-K. Kim, and S.~Malassiotis, ``Folding clothes autonomously: A complete pipeline,'' \emph{IEEE Transactions on Robotics}, vol.~32, no.~6, pp. 1461--1478, 2016.

\bibitem{weng2022fabricflownet}
T.~Weng, S.~M. Bajracharya, Y.~Wang, K.~Agrawal, and D.~Held, ``Fabricflownet: Bimanual cloth manipulation with a flow-based policy,'' in \emph{Conference on Robot Learning}.\hskip 1em plus 0.5em minus 0.4em\relax PMLR, 2022, pp. 192--202.

\bibitem{ha2022flingbot}
H.~Ha and S.~Song, ``Flingbot: The unreasonable effectiveness of dynamic manipulation for cloth unfolding,'' in \emph{Conference on Robot Learning}.\hskip 1em plus 0.5em minus 0.4em\relax PMLR, 2022, pp. 24--33.

\bibitem{sunil2023visuotactile}
N.~Sunil, S.~Wang, Y.~She, E.~Adelson, and A.~R. Garcia, ``Visuotactile affordances for cloth manipulation with local control,'' in \emph{Conference on Robot Learning}.\hskip 1em plus 0.5em minus 0.4em\relax PMLR, 2023, pp. 1596--1606.

\bibitem{sugiura2010foldy}
Y.~Sugiura, D.~Sakamoto, T.~A. Gowon, D.~Takahashi, M.~Inami, and T.~Igarashi, ``Foldy: Graphical teaching for garment-folding robot,'' in \emph{Workshop on Interactive System and Software}, 2010, pp. 7--12.

\bibitem{von2017naist}
F.~Von~Drigalski, D.~Yoshioka, W.~Yamazaki, S.-G. Cho, M.~Gall, P.~M.~U. Eljuri, V.~Hoerig, M.~Ding, J.~Takamatsu, T.~Ogasawara, \emph{et~al.}, ``Naist openhand m2s: A versatile two-finger gripper adapted for pulling and tucking textile,'' in \emph{2017 First IEEE International Conference on Robotic Computing (IRC)}.\hskip 1em plus 0.5em minus 0.4em\relax IEEE, 2017, pp. 117--122.

\bibitem{donaire2020versatile}
S.~Donaire, J.~Borras, G.~Alenya, and C.~Torras, ``A versatile gripper for cloth manipulation,'' \emph{IEEE Robotics and Automation Letters}, vol.~5, no.~4, pp. 6520--6527, 2020.

\bibitem{spiers2019using}
A.~J. Spiers, A.~S. Morgan, K.~Srinivasan, B.~Calli, and A.~M. Dollar, ``Using a variable-friction robot hand to determine proprioceptive features for object classification during within-hand-manipulation,'' \emph{IEEE Transactions on Haptics}, vol.~13, no.~3, pp. 600--610, 2019.

\bibitem{8989813}
Q.~Lu, A.~B. Clark, M.~Shen, and N.~Rojas, ``An origami-inspired variable friction surface for increasing the dexterity of robotic grippers,'' \emph{IEEE Robotics and Automation Letters}, vol.~5, no.~2, pp. 2538--2545, 2020.

\bibitem{borras2020grasping}
J.~Borras, G.~Alenya, and C.~Torras, ``A grasping-centered analysis for cloth manipulation,'' \emph{IEEE Transactions on Robotics}, vol.~36, no.~3, pp. 924--936, 2020.

\bibitem{kondratas2005robotic}
A.~Kondratas, ``Robotic gripping device for garment handling operations and its adaptive control,'' \emph{Fibres and Textiles in Eastern Europe}, vol.~13, no.~4, p.~84, 2005.

\bibitem{cubric2019study}
G.~Cubric and I.~Salopek~Cubric, ``Study of grippers in automatic handling of nonwoven material,'' \emph{Journal of The Institution of Engineers (India): Series E}, vol. 100, pp. 167--173, 2019.

\bibitem{thuy2013development}
M.~J. Thuy-Hong-Loan~Le, A.~Landini, M.~Zoppi, D.~Zlatanov, and R.~Molfino, ``On the development of a specialized flexible gripper for garment handling,'' \emph{Journal of automation and control engineering}, vol.~1, no.~3, 2013.

\bibitem{dragusanu2022dressgripper}
M.~Dragusanu, S.~Marullo, M.~Malvezzi, G.~M. Achilli, M.~C. Valigi, D.~Prattichizzo, and G.~Salvietti, ``The dressgripper: A collaborative gripper with electromagnetic fingertips for dressing assistance,'' \emph{IEEE Robotics and Automation Letters}, vol.~7, no.~3, pp. 7479--7486, 2022.

\bibitem{ono2007better}
E.~Ono and K.~Takase, ``On better pushing for picking a piece of fabric from layers,'' in \emph{2007 IEEE International Conference on Robotics and biomimetics (ROBIO)}.\hskip 1em plus 0.5em minus 0.4em\relax IEEE, 2007, pp. 589--594.

\bibitem{koustoumpardis2014underactuated}
P.~N. Koustoumpardis, K.~X. Nastos, and N.~A. Aspragathos, ``Underactuated 3-finger robotic gripper for grasping fabrics,'' in \emph{2014 23rd International Conference on Robotics in Alpe-Adria-Danube Region (RAAD)}.\hskip 1em plus 0.5em minus 0.4em\relax IEEE, 2014, pp. 1--8.

\bibitem{shibata2008handling}
M.~Shibata, T.~Ota, Y.~Endo, and S.~Hirai, ``Handling of hemmed fabrics by a single-armed robot,'' in \emph{2008 IEEE International Conference on Automation Science and Engineering}.\hskip 1em plus 0.5em minus 0.4em\relax IEEE, 2008, pp. 882--887.

\bibitem{li2015folding}
Y.~Li, Y.~Yue, D.~Xu, E.~Grinspun, and P.~K. Allen, ``Folding deformable objects using predictive simulation and trajectory optimization,'' in \emph{2015 IEEE/RSJ International Conference on Intelligent Robots and Systems (IROS)}.\hskip 1em plus 0.5em minus 0.4em\relax IEEE, 2015, pp. 6000--6006.

\bibitem{jia2018manipulating}
B.~Jia, Z.~Hu, J.~Pan, and D.~Manocha, ``Manipulating highly deformable materials using a visual feedback dictionary,'' in \emph{2018 IEEE International Conference on Robotics and Automation (ICRA)}.\hskip 1em plus 0.5em minus 0.4em\relax IEEE, 2018, pp. 239--246.

\bibitem{lee2015learning}
A.~X. Lee, H.~Lu, A.~Gupta, S.~Levine, and P.~Abbeel, ``Learning force-based manipulation of deformable objects from multiple demonstrations,'' in \emph{2015 IEEE International Conference on Robotics and Automation (ICRA)}.\hskip 1em plus 0.5em minus 0.4em\relax IEEE, 2015, pp. 177--184.

\bibitem{cuen2008action}
S.~Cu{\'e}n~Roch{\'\i}n, J.~Andrade-Cetto, and C.~Torras, ``Action selection for robotic manipulation of deformable objects,'' 2008.

\bibitem{maitin2010cloth}
J.~Maitin-Shepard, M.~Cusumano-Towner, J.~Lei, and P.~Abbeel, ``Cloth grasp point detection based on multiple-view geometric cues with application to robotic towel folding,'' in \emph{2010 IEEE International Conference on Robotics and Automation}.\hskip 1em plus 0.5em minus 0.4em\relax IEEE, 2010, pp. 2308--2315.

\bibitem{willimon2011model}
B.~Willimon, S.~Birchfield, and I.~Walker, ``Model for unfolding laundry using interactive perception,'' in \emph{2011 IEEE/RSJ International Conference on Intelligent Robots and Systems}.\hskip 1em plus 0.5em minus 0.4em\relax IEEE, 2011, pp. 4871--4876.

\bibitem{sun2015accurate}
L.~Sun, G.~Aragon-Camarasa, S.~Rogers, and J.~P. Siebert, ``Accurate garment surface analysis using an active stereo robot head with application to dual-arm flattening,'' in \emph{2015 IEEE international conference on robotics and automation (ICRA)}.\hskip 1em plus 0.5em minus 0.4em\relax IEEE, 2015, pp. 185--192.

\bibitem{garcia2022household}
I.~Garcia-Camacho, J.~Borr{\`a}s, B.~Calli, A.~Norton, and G.~Aleny{\`a}, ``Household cloth object set: Fostering benchmarking in deformable object manipulation,'' \emph{IEEE Robotics and Automation Letters}, vol.~7, no.~3, pp. 5866--5873, 2022.

\bibitem{8411094}
A.~J. Spiers, B.~Calli, and A.~M. Dollar, ``Variable-friction finger surfaces to enable within-hand manipulation via gripping and sliding,'' \emph{IEEE Robotics and Automation Letters}, vol.~3, no.~4, pp. 4116--4123, 2018.

\bibitem{clark2023household}
A.~B. Clark, L.~Cramphorn-Neal, M.~Rachowiecki, and A.~Gregg-Smith, ``Household clothing set and benchmarks for characterising end-effector cloth manipulation,'' in \emph{2023 IEEE International Conference on Robotics and Automation (ICRA)}.\hskip 1em plus 0.5em minus 0.4em\relax IEEE, 2023, pp. 9211--9217.

\bibitem{chitta2012moveit}
S.~Chitta, I.~Sucan, and S.~Cousins, ``Moveit![ros topics],'' \emph{IEEE Robotics \& Automation Magazine}, vol.~19, no.~1, pp. 18--19, 2012.

\bibitem{hoque2022learning}
R.~Hoque, K.~Shivakumar, S.~Aeron, G.~Deza, A.~Ganapathi, A.~Wong, J.~Lee, A.~Zeng, V.~Vanhoucke, and K.~Goldberg, ``Learning to fold real garments with one arm: A case study in cloud-based robotics research,'' in \emph{2022 IEEE/RSJ International Conference on Intelligent Robots and Systems (IROS)}.\hskip 1em plus 0.5em minus 0.4em\relax IEEE, 2022, pp. 251--257.

\bibitem{canny1986computational}
J.~Canny, ``A computational approach to edge detection,'' \emph{IEEE Transactions on pattern analysis and machine intelligence}, no.~6, pp. 679--698, 1986.

\end{thebibliography}

\end{document}